\documentclass[letterpaper, 10 pt, journal, twoside]{ieeetran}  % Comment this line out
\usepackage{amssymb,graphicx,amsmath,color,algorithm,algorithmic,url}%,tabularx,subcaption,float}
\usepackage[english]{babel}
\usepackage{blindtext}
\usepackage{afterpage}
\usepackage[noadjust]{cite} % groups citations
\usepackage{hhline,caption}
\usepackage{hyperref}
\usepackage{diagbox} 
\usepackage{multirow}
\usepackage[font=small,labelfont=bf]{caption}
\IEEEoverridecommandlockouts                              % This command is only
\title{
Sim-to-Real Model-Based and Model-Free Deep Reinforcement Learning for Tactile Pushing
}
\author{
Max Yang, \quad
Yijiong Lin, \quad
Alex Church,\quad
John Lloyd, \quad
Dandan Zhang, \\
David A.W. Barton$^*$,\quad
Nathan F. Lepora$^*$ \\% <-this % stops a space

\thanks{This work was supported by the EPSRC Doctoral Training Partnership (DTP) scholarship. NL was supported by an award from the Leverhulme Trust on ‘A biomimetic forebrain for robot touch’ (RL-2016-39). (\textit{Corresponding author: Max Yang})}
\thanks{All authors are with the Department of Engineering Mathematics and Bristol Robotics Laboratory, University of Bristol, Bristol BS8 1UB, U.K. (email: \{max.yang, yijiong.lin, alex.church, john.lloyd, ye21623, david.barton, n.lepora\}@bristol.ac.uk)}
\thanks{$^{*}$ These authors provided equal supervision.}
}

\begin{document}
\maketitle
% \pagestyle{empty}
% \thispagestyle{empty}
% Comment or remove these lines for final RAL version.
%----------------------------------------------------------------------------------
%\bstctlcite{IEEEexample:BSTcontrol}
%----------------------------------------------------------------------------------
%----------------------------------------------------------------------------------
\begin{abstract}
%----------------------------------------------------------------------------------

% --------------------- Submission Version -------------------
Object pushing presents a key non-prehensile manipulation problem that is illustrative of more complex robotic manipulation tasks. While deep reinforcement learning (RL) methods have demonstrated impressive learning capabilities using visual input, a lack of tactile sensing limits their capability for fine and reliable control during manipulation. Here we propose a deep RL approach to object pushing using tactile sensing without visual input, namely tactile pushing. We present a goal-conditioned formulation that allows both model-free and model-based RL to obtain accurate policies for pushing an object to a goal. To achieve real-world performance, we adopt a sim-to-real approach. Our results demonstrate that it is possible to train on a single object and a limited sample of goals to produce precise and reliable policies that can generalize to a variety of unseen objects and pushing scenarios without domain randomization. We experiment with the trained agents in harsh pushing conditions, and show that with significantly more training samples, a model-free policy can outperform a model-based planner, generating shorter and more reliable pushing trajectories despite large disturbances. The simplicity of our training environment and effective real-world performance highlights the value of rich tactile information for fine manipulation. Code and videos are available at \url{https://sites.google.com/view/tactile-rl-pushing/}.
% --------------------- Submission Version -------------------

% Code and videos are available at \url{https://github.com/ac-93/tactile_gym} and \url{https://sites.google.com/my.bristol.ac.uk/tactilegym2}. 

\end{abstract}

%----------------------------------------------------------------------------------
% Keywords appear just beneath the abstract. Use only for final RAL version. 

%----------------------------------------------------------------------------------
\section{INTRODUCTION}\label{sec:Intro}
%----------------------------------------------------------------------------------

\begin{figure}[ht]
    \centering  
    \includegraphics[width=0.98\linewidth]{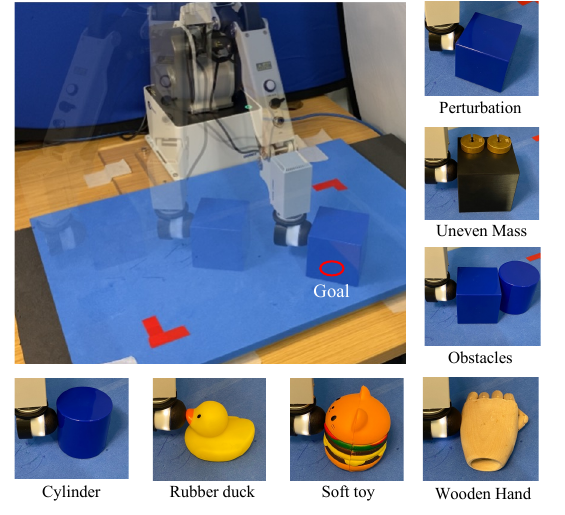}
    \vspace{-0.5em}
    \caption{Real-world object pushing setup, comprising a desktop robot (Dobot MG400) mounted on a pushing platform with a Tactip attached to serve as the pusher.}
    \vspace{-1.0em}
    \label{fig:first_image}
\end{figure}

%-------------------------------------------------------- 

% -------------------- Submission Version: Object pushing  (unchanged)---------------------
\IEEEPARstart{D}{eveloping} a system for general object pushing remains an unsolved challenge due to the partially observable system and difficult-to-model physics. Prior studies have used approximated analytical models \cite{hogan2016feedback}, data-driven models \cite{bauza2018data}, or a hybrid approach \cite{kloss2022combining} to obtain controllers for pushing. Using analytical methods can offer performance guarantees, but they are often limited by the underlying assumptions. Data-driven techniques are more effective in overcoming these limitations but they require large datasets of real-world pushing \cite{yu2016more, bauza2019omnipush} and can fail to generalize to new situations. As the demand for versatile controllers for robot manipulation increases, reinforcement learning (RL) has become an attractive option due to its generality and ability to model complex relationships. However, most RL studies on pushing have relied on vision-based systems, either for direct policy training~\cite{zeng2018learning} or for providing object-centric information \cite{manuelli2020keypoints}. These systems can suffer from low accuracy and occlusions \cite{krivic2019pushing}. 
% -------------------- Submission Version: Object pushing (unchanged) ---------------------

% ------------------- Submission Version: Tactile pushing (unchanged) ---------------------

Tactile sensing captures detailed contact information during robot-object interaction that enables precise control of contacts. Early work by Lynch et al. \cite{lynch1992manipulation} showed the prospects of tactile feedback for translating objects to target orientations. Recently, Lloyd et al. \cite{lloyd2021goal} adopted a state-feedback control approach to achieve goal-driven pushing using a trajectory composed of discrete taps. Their works suggest that accurate pushing of objects with varying physical properties is achievable using only tactile and proprioceptive states as input, and illustrate the potential generalization benefits of tactile input. However, traditional feedback control methods can be limited, especially when the problem becomes highly nonlinear. 

% ------------------- Submission Version: Tactile pushing (unchanged) ---------------------

% ---------------- Submission Version: Motivation and contribution ---------------------
In this study, we explore deep RL approaches for tactile pushing, which can provide more general and versatile methods that do not require the explicit design of paramterised controllers. A similar problem was studied in \cite{church_tactile_2021, lin2022tactile} where a model-free agent learned to push objects through a predetermined path of goals with short-distance intervals. Instead, we examine a more difficult problem of pushing an object to an arbitrarily placed distant goal, which offers a greater degree of variability in the goals that the agent must achieve. This allows the agent to search for the shortest path, resulting in a more versatile and adaptable pushing solution. 

The success of training RL policies relies heavily on using features that are important for the task. When using optical tactile sensors, instead of learning directly from tactile images, as in \cite{church_tactile_2021, lin2022tactile}, we use pose-based observations which provide specific pushing-related contact features that can be more efficient for learning. These contact features are derived from tactile images, and we refer to them as tactile pose.

Specifically, we present a formulation that allows model-free and model-based RL to solve an object-pushing problem reliably without visual input. In the real-world setup, we use the TacTip \cite{ward2018tactip}, a soft hemispherical optical tactile sensor that provides contact information through pin motion under its sensing surface. Prior work has shown the TacTip's ability to accurately predict the surface pose of the contacted object \cite{lepora2020optimal}. We leverage the sensor jointly with the tactile sim-to-real framework developed by Church et al. \cite{church_tactile_2021} to achieve real-world tactile pushing. Our approach requires training only in simulation to push a cube but can generalize to unknown objects with different physical properties and remain robust to disturbances. To the best of our knowledge, this is the first successful application of model-based RL within a sim-to-real framework. We investigate this further by comparing the online performance of a model-free policy learned offline (model-free RL) against an online planner that plans with a learned model (model-based RL) for this task. 

The main contributions of this paper are:\\
\noindent 1) We formulate the tactile pushing task as a goal-conditioned RL problem and obtain reliable policies in simulation for reaching arbitrary distant goals.\\ 
\noindent 2) We present a sim-to-real RL pipeline for pose-based tactile observations and demonstrate the benefits of using the contact surface pose as tactile input for pushing.\\
\noindent 3) We perform an empirical study on the generalizability and robustness of the final policies obtained from model-free and model-based RL by testing on challenging unseen pushing scenarios in both simulation and the real-world environment. 

% mitigating the need for accurate modeling of different objects or a diverse training curriculum involving multiple objects. 
% ---------------- Submission Version: Motivation and contribution ---------------------

%----------------------------------------------------------------------------------
\section{RELATED WORK} \label{sec:related}
%----------------------------------------------------------------------------------

% ---------------- Submission Version: Related Work ---------------------

Early work on planar pushing focused on developing analytical models from first principles. Mason et al. \cite{mason1986mechanics} introduced the voting theorem for determining the direction of rotation of a pushed object. Lynch et al. \cite{lynch1992manipulation} used the concept of limiting surface to translate and orientate objects using only tactile feedback. More recent works have attempted to overcome the limitations of analytical approaches through data-driven methods. Both Kloss et al. \cite{kloss2022combining} and Ajay et al. \cite{ajay2018augmenting} used hybrid approaches by combining an analytical model with neural networks learned using MIT Push Dataset \cite{yu2016more} to improve the modelling of pushing dynamics. Given a predictive model, Model Predictive Control (MPC) has become a popular feedback control method for complex contact dynamics~\cite{hogan2016feedback}. Bauza et al. \cite{bauza2018data} and Arruda et al. \cite{arruda2017uncertainty} used similar ideas to fit a Gaussian Process model from real-world pushing data, whilst Cong et al. \cite{cong2020self} fitted an LSTM-based dynamics model using pushing data collected by an expert policy in simulation with domain randomization. Manuelli et al. \cite{manuelli2020keypoints} developed a visual-based dynamics model for planning using object key points and a small dataset of random interactions. 

In contrast to model-based methods, model-free control offers an alternative perspective on the planar pushing task. Krivic et al. \cite{krivic2019pushing} developed an adaptive feedforward/feedback controller for mobile robot pushing. However, the success of the proposed method relied on the accuracy of the vision-based measurements. Lloyd et al. \cite{lloyd2021goal} developed a state feedback controller with tactile input that achieved accurate and stable object pushing. However, the proposed controller was shown to be limited when dealing with complex dynamics very near the goal. With deep RL, pushing has become a popular benchmark task for robot manipulation \cite{plappert2018multi}. Researchers have demonstrated impressive results from using deep RL applied to vision-based systems \cite{ebert2018visual, nair2018visual, cong2022reinforcement}. However, they commonly neglect valuable contact information available from tactile sensing. Incorporating tactile into deep RL has been shown to be advantageous for generalizing to novel objects \cite{dong2021tactile}, which is important for robot manipulation. Learning from tactile observations has been made more accessible through simulations such as \cite{Ding2020Sim-to-RealSensing, bauza2020tactile, gomes2021generation, wang2022tacto, si2022taxim} which can accelerate tactile RL training without relying solely on real-world interactions. For this reason, we leverage Tactile Gym \cite{church_tactile_2021} and a sim-to-real framework to achieve our task. 

% ---------------- Submission Version: Related Work ---------------------

%----------------------------------------------------------------------------------
\section{Method} \label{sec:method}
%----------------------------------------------------------------------------------

% %--------------------------------------------------------    
\begin{figure*}[t]
    \centering
    \includegraphics[width=1\linewidth]{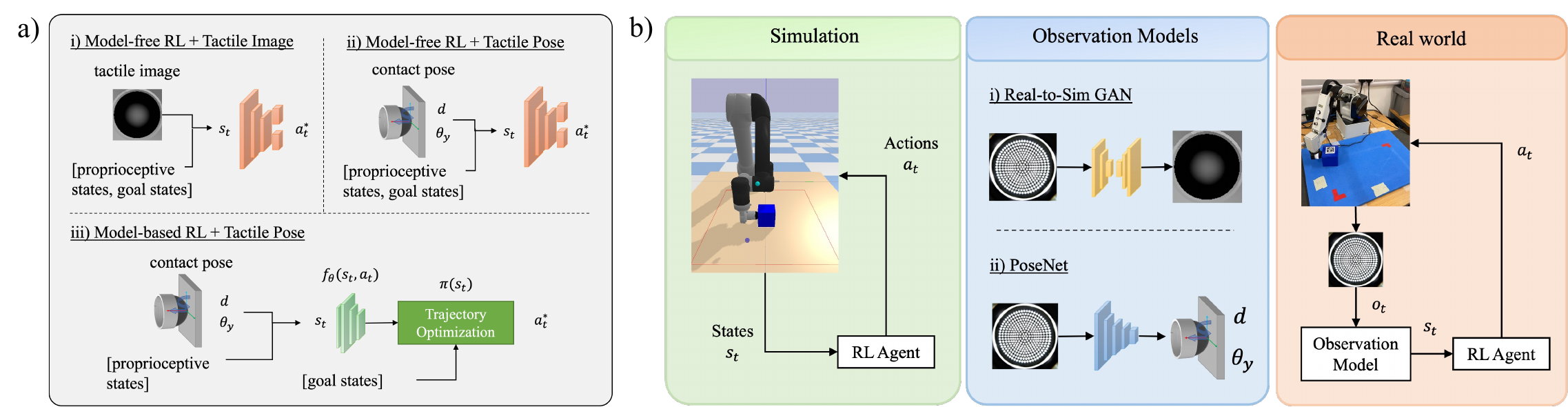}
    \caption{Overview of modelling and sim-to-real transfer. (a) The three tactile RL pipelines: i) a model-free agent trained on tactile-image-based observations, ii) a model-free agent trained on pose-based observations, and iii) a model-based agent trained on pose-based observations. (b) Workflow for sim-to-real tactile RL: the RL policy is trained with simulated tactile observations; tactile data is then collected to train an observation model to bridge the sim-to-real gap; then the RL policy and observation model are combined for real-world implementation.}
    \label{fig:workflow}
    \vspace{-0.75em}
\end{figure*}
% %---------

% ---------------- Submission Version: Problem Formulation ---------------------

% %--------------------------------------------------------

%----------------------------------------------------------------------------------
\subsection{Reinforcement Learning} \label{subsec:rl_formulation}
%----------------------------------------------------------------------------------

We formulate the task as a finite horizon Markov Decision Process (MDP) defined by a continuous state $s \in \mathcal{S}$, a continuous action space $a \in \mathcal{A}$, a probabilistic state transition function $p(s_{t+1}|s_t, a_t)$, and a reward function $r \in \mathcal{R} : \mathcal{S}\times\mathcal{A} \times\mathcal{G}\xrightarrow{} \mathbb{R}$. To investigate the two paradigms of deep RL, we use a model-free algorithm that learns a policy directly and a model-based algorithm that plans using a learned model, to solve this MDP.

\subsubsection{\textbf{Model-free RL}}
To achieve multiple goals with a unified policy, we augment the original MDP with a goal-conditioned one \cite{kaelbling1993learning}. The aim is to obtain a policy $\pi_{\theta}^*(a_t|s_t, g)$ parameterized by $\theta$ that maximizes the expected return over an episode $\tau$ and goals $g$ given by:
\begin{equation}
   \!\pi_\theta^* = \arg \max_{\pi} \mathbb{E}_{\tau \sim p_{\pi} (\tau), \, g  \sim q (g) }  \left[ \sum_{t=0}^{T} \gamma^t r(s_t, a_t, g) \right] \!,
   \label{eq:expect_rew}
\end{equation}
where $\gamma \in [0, 1)$ is the discount factor, $p_{\pi}(\tau)$ is a distribution over episodes conditioned on $\pi$, $q (g)$ is a distribution over the goal space $\mathcal{G}$ with $g \in \mathcal{G}$. A goal is randomly sampled at the start of the episode and remains fixed until the episode ends. At each time step, the policy receives goal-aware observations $(s_{t}, g)$ and obtains a reward $r(s_t, a_t, g)$ given an action $a_t$ provided by policy $\pi: \mathcal{S} \times \mathcal{G} \rightarrow \mathcal{A}$. For our task, we train using an off-policy algorithm, Soft Actor-Critic (SAC) \cite{haarnoja2018soft} implemented using stable-baselines3 \cite{raffin2019stable}.

\subsubsection{\textbf{Model-Based RL with Online Planning}}
The objective of the RL agent is to learn a reliable dynamics model to be used for forward predictions. We use probabilistic ensemble neural networks $f_{\theta}$ parameterized by $\theta$ to represent the dynamics model which approximates the probabilistic state transition function $p(s_{t+1}|s_t, a_t)$. The predictive model returns a Gaussian distribution with diagonal covariances $f_\theta(s_t, a_t) \equiv \mathcal{N}(\mu_\theta(s_t, a_t), \Sigma_\theta(s_t, a_t))$, modelling the change in states $\Delta s_t = s_{t+1} - s_t$ where $\Delta s_t \sim f_\theta(s_t, a_t)$.  Given a set of state transitions $\mathcal{D} = \{(s_t, a_t, s_{t+1})_{1,...N}\}$, we train the model on a 1-step negative log-likelihood loss (NLL):
\begin{equation}
    \mathcal{L}_\text{NLL} = \sum_{n=1}^{N}[s_{n+1} - \mu_\theta]^T \Sigma_\theta^{-1}[s_{n+1} - \mu_\theta] + \text{log}( \text{det} ( \Sigma_\theta)).
\end{equation}
We use the learned model for planning in an MPC framework. At each time step $t$, planning solves a trajectory optimization problem: $\arg \max_{a_{k:t+H}} \mathbb{E}_f\left[\sum_{k=t}^{t+H} r(s_k, a_k, g)\right]$ to find the optimal action sequence $a_{k:t+H}$ for a prediction horizon $H$ at the current state $s_t$. The first action of this action sequence is then applied before re-planning again in the next time step. 

During RL training, the model training and data collection are interleaved whereby trajectories produced by the planner are stored to train the model periodically. This way, the model and planning improve simultaneously until convergence (see Ref.~\cite{chua2018deep} for more details). We use Probabilistic Ensemble Trajectory Sampling (PETS) implemented on MBRL-LIB \cite{pineda2021mbrl}. We experimented with cross-entropy method (CEM) and model-predictive path integral (MPPI) as the MPC optimizer and found that CEM performed best for our application.

%----------------------------------------------------------------------------------
\subsection{Task Formulation}\label{subsec:task_formulation}
%----------------------------------------------------------------------------------

We consider the task of pushing an unknown object to any randomly sampled goal position inside a predefined workspace from an initial contact. By relying solely on tactile sensing to gather object-centric information, this task is challenging due to the lack of explicit information about the object properties, e.g. its centre. To circumvent this issue, we formulate the objective into one where we control the object contact pose relative to the pusher instead. We ensure controllability by maintaining continuous contact with the object and we encourage stable pushing motions by keeping the object's contact surface normal to the pusher. In this section, we formulate these conditions into an RL objective that can achieve stable pushing towards an arbitrarily placed goal.

%-------------------Submission Version: tactile observation -----------------------------
%----------------------------------------------------------------------------------

\subsubsection{\textbf{Tactile Observations}} \label{subsubsec:observations}
Whilst various tactile observations can be used to accomplish the same task, the learned policies can have varying performances. On one hand, tactile images provide a general way to encode detailed contact features; on the other hand, tactile-derived contact surface pose (referred to as tactile pose) can be more effective at representing pushing-related features. We construct two goal-aware observations ($\mathcal{S}_1$ and $ \mathcal{S}_2$) for model-free RL using these two types of tactile input and compare their effectiveness. We also use the contact surface pose to construct a pose-based state space ($\mathcal{S}_3$) for the model-based agent to learn a dynamics model. 

In the following observation definitions, the notation $s^a_b$ denotes the relative value of the state variable of $b$ w.r.t. $a$, and $s_b$ is the value of the state variable of $b$. The letter $p$ represents the robot pusher, $o$ is the object contact surface, and $g$ is the goal:
\begin{align}\label{eq:tactile_obs}
  \mathcal{S}_1 &= [I_{\text{tactile}}, \,\, x^p_{g}, \,\,y^p_{g}, \,\,\theta^p_{g}], \notag\\
  \mathcal{S}_2 &= [{x}^p_o, \,\, y^p_{o}, \,\, \theta^p_{o}, \,\, x^o_{g}, \,\, y^o_{g}, \,\, \theta^o_{g}],\\
  \mathcal{S}_3 &= [x^p_{o}, \,\,y^p_{o}, \,\,\theta^p_{o}, \,\,x_{o}, \,\,y_{o}, \,\,\theta_{o}], \notag 
\end{align}
where the state variables $x$, $y$ represent the position, $\theta$ the orientation, and $I_{\text{tactile}}$ the tactile image. $\mathcal{S}_1$ is the tactile-image-based goal-aware observation, $\mathcal{S}_2$ is the tactile-pose-based goal-aware observation, $\mathcal{S}_3$ is the pose-based state space. Figure \ref{fig:workflow}a) demonstrates the construction of different input observations and their integration with different RL methods. Accordingly, we train three corresponding RL agents with tactile input for the pushing task. 

%---------------Submission Version: tactile observation -----------------------------
%----------------------------------------------------------------------------------

%---------------Submission Version: Action space -----------------------------
%----------------------------------------------------------------------------------

\subsubsection{\textbf{Action Space}} 
To control the robot pusher, we use position control and define the action space as $\Delta y \in [\text{-1mm}, \text{1mm}]$ and $\Delta \theta \in [-1^\circ, 1^\circ]$, representing the change in position and orientation at each time step in the pusher's frame of reference. To reduce the dimensionality of the problem, we fix the forward position control of the pusher to have a constant $\Delta x=\text{1mm}$ action. This is a reasonable assumption since we want the robot to push the object without losing contact. With this simplification, some goals may be unreachable, particularly when the goals are very close to the object. However, our experimental results will show that this issue affects a limited number of cases and is an acceptable trade-off for greater robustness gains. 

%---------------Submission Version: Action space -----------------------------
%----------------------------------------------------------------------------------

\subsubsection{\textbf{Reward Shaping}}

%---------------Submission Version: Reward shapping -----------------------------
%----------------------------------------------------------------------------------

Long-horizon goal-conditioned problems are known to be challenging for RL. Standard reward formulations can result in overly greedy policies that can fail to complete the task. Sparse reward presents a difficult exploration problem that is not effective for planning. A simple dense reward composed of the distance between the achieved and desired goals can lead to local optima where the agent may first need to increase the distance to the goal (e.g. turning the object around) before reaching it \cite{trott2019keeping}. 

For this problem, we take a reward-shaping approach. To do so, we additionally define a desired contact orientation as the object-goal bearing angle, $g_\theta = \mathrm{atan2}(o_{xy}, g_{xy})$, to guide the pushing direction towards the goal. We use the distance functions $f(a, b)$ and $g(a, b)$ to represent the Euclidean and cosine distances between $a$ and $b$, respectively. Our reward function contains components $f(o_{xy}, g_{xy})$, the Euclidean distance between the object contact position and goal position, $g(o_\theta, g_\theta)$, the cosine distance between the contact surface orientation and the desired orientation, and $g(p_\theta,o_\theta)$, the cosine distance between the pusher orientation and the contact surface orientation:
\begin{equation}\label{eq:push_reward}
\!r =
    \begin{cases}
      - (g(o_\theta,g_\theta) + g(p_\theta,o_\theta)) & \text{if $||o_{xy} - g_{xy}|| > d$},\\
      - (f(o_{xy}, g_{xy}) + g(p_\theta,o_\theta)) & \text{if $||o_{xy} - g_{xy}|| \leq d$}.\\
    \end{cases}       
\end{equation}

The first part of this reward function underlies the goal-driven pushing objective: given a constant forward velocity, the optimal pushing path can be achieved by orientating the object towards the goal, represented by the reward signal $g(o_\theta, g_\theta)$. 

In theory, this should be sufficient to complete the task, but in practice, we found that it caused undesirable behaviors near the goal: a phenomenon that arises because the goal-bearing angle reward signal $g_\theta$ is not effective near the goal location \cite{lloyd2021goal}. To alleviate this issue, in the second part of the reward function, we use a Euclidean-distance-based reward signal $f(o_{xy}, g_{xy})$ when the object is inside an approaching zone of distance $d$ from the goal (here using $d=100$\,mm). 

To ensure the goal can be reached reliably, we augment both parts of the reward with an additional signal $g(p_{\theta}, o_{\theta})$ that encourages the agent to maintain a normal contact pose. This guides the pushing direction to be normal to the object contact surface. We found that this component was critical for successful training, as it encourages the pusher to push through the center of friction (COF) \cite{lynch1992manipulation}, with multiple benefits; first, it encourages the pusher to maintain contact with the object, especially when the goal bearing angle is large and the agent likely to lose contact; second, if normality of pushing holds true, then as the contact location moves towards the goal, the COF also moves towards the goal; and third, stable pushing also improves model learning to better predict the dynamics and avoid compounded errors becoming problematic for action selection. This reward function provides a dense reward for efficient learning whilst also avoiding problems with local optima. We found that any missing components from this reward design caused pushing instabilities and failure to learn.

%---------------Submission Version: Reward shapping -----------------------------
%----------------------------------------------------------------------------------

\subsubsection{\textbf{Goal space, resets and termination}} 

% ----------------------------------------------------------------------------------

\begin{figure}[t]
  \centering
%  \hfill\begin{minipage}{0.45\linewidth}
     \includegraphics[width=1\linewidth]{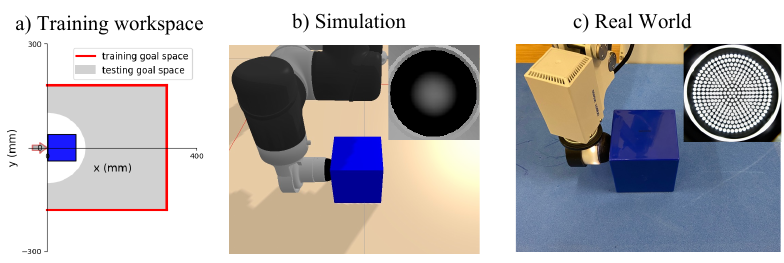}
     \vspace{-1em}
    \caption{Diagrams of training and testing environments. The training goal space is set near the edges of the robot workspace, then we test for goals across the entire workspace.}
  \label{fig:training_workspace}
  \vspace{-.5em}
\end{figure}

% ----------------------------------------------------------------------------------

% %---------------Submission Version: Goal space, resets and termination ---------------------------
% %----------------------------------------------------------------------------------

The workspace is a rectangle of extent \mbox{$x\in[0, 400]$\,mm} and \mbox{$y\in\![-300, 300]$\,mm}.  The goal space is uniformly distributed near the edge of this workspace (see Fig.~\ref{fig:training_workspace}a) and the agent is initialized in contact with the object at the origin in a stable push configuration. This training regime avoids sampling goals that are unreachable and ensures sufficient exploration of the workspace. As we will show later, the limited goal sampling does not impede the performance of the final policy and reaching goals outside of this goal space can still be achieved. The agents have access to the contact location and goal location. Each episode is terminated when the contact is within 25\,mm of the goal. We found this tolerance was sufficient for the object sizes that are considered in our problem.

% %---------------Submission Version: Goal space, resets and termination ---------------------------
% %----------------------------------------------------------------------------------

%-------------------------------------------------------- 
\subsection{Sim-to-Real Transfer}
%-------------------------------------------------------- 
For real-world deep RL with tactile information, we adopt the sim-to-real framework of \cite{church_tactile_2021}, comprising 3 main steps: 1)~training the agent in simulation, 2) collecting tactile data and training observation models to bridge the sim-to-real gap, and 3) performing zero-shot sim-to-real policy transfer. As the framework was designed for tactile image observations, we achieve sim-to-real for pose-based tactile observations by adapting it with a PoseNet \cite{lepora2020optimal} where step 2) is replaced with pose training (workflow of sim-to-real procedure in Fig.~\ref{fig:workflow}b). 

%-------------------------------------------------------- 
\subsubsection{Simulation with Tactile Information}
%-------------------------------------------------------- 
The Tactile Gym \cite{church_tactile_2021,lin2022tactile}  used here relies on PyBullet to simulate rigid-body physics. To obtain the tactile observations specified in Eq.~\ref{eq:tactile_obs}, the rendered depth image of the simulated tactile sensor is used for the tactile image and the contact information between the approximated tactile sensor and object is used to infer contact surface pose. The training objective is to push a cube (edge length 75\,mm) towards randomly sampled goals in the goal space (see Fig.~\ref{fig:training_workspace}b). Both model-free and model-based RL are trained from scratch without domain randomisation and we use default physics parameters set by the original Tactile Gym.

%-------------------------------------------------------- 
\subsubsection{Tactile Observation Models}
%-------------------------------------------------------- 
For tactile-image-based observations, we use a Generative Adversarial Network (GAN) to train an image translation model; for tactile-pose-based observations, we train a Convolutional Neural Network (CNN) to predict~pose. 

\textbf{Real-to-Sim GAN}: The model is used for domain adaptation, translating real tactile images into simulated tactile images. We use pix2pix for the generator, which has a U-net architecture, with a standard CNN for the discriminator. For training, the same data collection procedure is performed in simulation and reality to generate a dataset of real and simulated image pairs; then the image-to-image translation is achieved via supervised learning (see Ref.~\cite{church_tactile_2021} for details).

%---------------Submission Version: Goal space, resets and termination -----------------------------
%----------------------------------------------------------------------------------

\textbf{PoseNet}: Here we use a CNN model as a tactile sim-to-real transfer method via feature extraction. Given a real tactile image, the model estimates the relative contact pose between the object surface and sensor. For 2D surfaces, these pose variables are contact depth and angle (polar coordinates). This is then converted into Cartesian coordinates to provide the tactile pose information used in Eq. \ref{eq:tactile_obs}. During data collection, robot poses relative to a fixed contact surface are stored as labels for training (see refs~\cite{lepora2020optimal,lepora2022digitac} for more details). 

\textbf{Data Collection}: For both models, we collected tactile-image data using the same sampling ranges to encode equivalent contact features (see Table \ref{table:ranges}). Specifically, we collected surface data by moving the sensor to randomly sampled poses on a flat surface of a fixed 3D-printed stimulus and stored those tactile images with their corresponding labels.  

%---------------Submission Version: Goal space, resets and termination -----------------------------
%----------------------------------------------------------------------------------

\begin{table}[h]
% shorten the width of each column
\addtolength{\tabcolsep}{-1pt}
\vspace{0em}
\caption{Sensor pose sampling ranges used during tactile data collection, relative to a fixed sensor coordinate frame.}
\centering
\begin{tabular}{c|c@{}c@{}} 
 & \textbf{    Tactile GAN    } & \textbf{    PoseNet    } \\ 
\hline
Depth range (mm) & [-1, -5] & [-1, -5]\\
Angle range (deg)& [-30, 30] & [-30, 30] \\
Train samples & 5000 & 1476 \\
Val samples & 2000 & 524 \\
Val accuracy & SSIM: 0.99 &  [$\pm0.05$mm, $\pm0.6$deg]
\end{tabular}
\label{table:ranges}
\end{table}

%-------------------------------------------------------- 
\subsubsection{Zero-shot Policy Transfer}
%-------------------------------------------------------- 
With the trained observation models, the policies trained using different simulated tactile observations in step 1) can now be used for real-world robotic tasks. For model-free RL, this involves transferring the learned policy to reality. For the model-based case, the dynamics model is transferred to reality, where the optimizer will solve the task in an online manner.

%--------------------------------------------------------    
\subsection{Hardware System Overview}\label{subsec:TS}
%--------------------------------------------------------    
For the real-world experiment, we use the robot system setup presented in \cite{lin2022tactile,lepora2022digitac}. This comprises a Dobot MG400 4-axis desktop robot arm, with maximum payload of 750\,g, maximum reach of 440\,mm and maximum repeatability $\pm0.05$\,mm. The robot has 4 degrees of freedom with end effector rotation just around the $z$-axis. For planar pushing, these degrees-of-freedom are ideally suited to the task. The end effector is equipped with a horizontally-mounted tactile sensor/pusher, for which we use a 331-pin TacTip. To validate pushing trajectories, we placed ArUco markers on top of the test objects and used a tracking method described in \cite{lloyd2021goal}.

%--------------------------------------------------------    
\section{Experiments and Results}\label{sec:exps_ressults}
%-------------------------------------------------------- 

%-------------------------------------------------------- 
\subsection{Training in Simulation}
%-------------------------------------------------------- 

% --------------------------------------------------------------------------------

\begin{figure}[t]
  \centering
%  \hfill\begin{minipage}{0.45\linewidth}
     \includegraphics[width=1\linewidth]{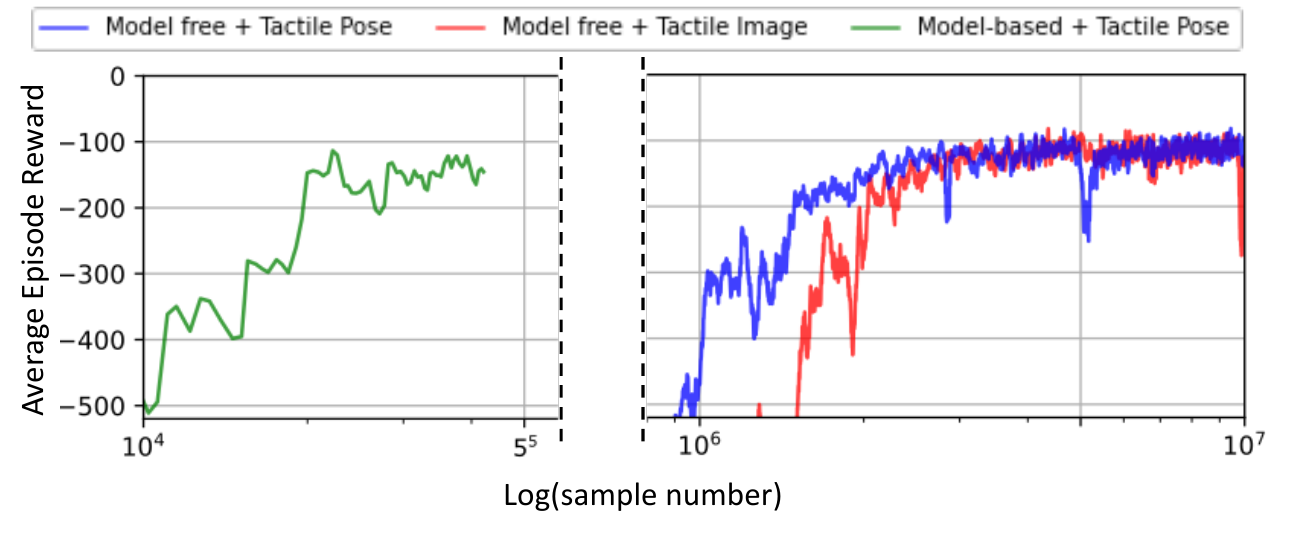}
     \vspace{-2em}
    \caption{Learning curves during training for model-based (left) and model-free RL (right) on the same log axis. }
  \label{fig:learning_curve}
\end{figure}

% ----------------------------------------------------------------------------------

The learning curves are shown in Fig. \ref{fig:learning_curve}. The agents are evaluated and the asymptotic performance is recorded in Table~\ref{table:learning_results}. As expected, the model-free agent needed significantly more samples ($100\times$) as compared to the model-based case. However, with significantly more training samples, both of the model-free agents achieved higher (less negative) best rewards.  We also note that in the model-free case, using tactile pose was more sample efficient and achieved a better reward than tactile images, which suggests contact surface pose is a more effective feature for learning. 

% This is consistent with earlier work where model exploitation using an MPC planner can limit the asymptotic performance of model-based RL \cite{janner2019trust}.

\begin{table}[h]
\vspace{0em}
\addtolength{\tabcolsep}{-1pt}
\caption{Training samples for within 10\% of best reward and final best rewards.}
\centering
\begin{tabular}{l|cc} 
\hline
\textbf{RL Agent} & \textbf{Samples} & \textbf{Best Reward} \\ 
\hline
\textbf{Model Free + Tactile Image}     & 3.2m          & -124.86    \\ 
\textbf{Model Free + Tactile Pose}         & 2.8m          & \textbf{-122.85}   \\
\textbf{Model Based + Tactile Pose}  & \textbf{25k} & -144.70  \\ 
\hline
\end{tabular}
\label{table:learning_results}
\end{table}

  %---------------------------------------------------
\subsection{Simulation Performance}
% %--------------------------------------------------------    

\begin{figure}[b]
\vspace{-1.5em}
  \centering
     \includegraphics[width=1\linewidth]{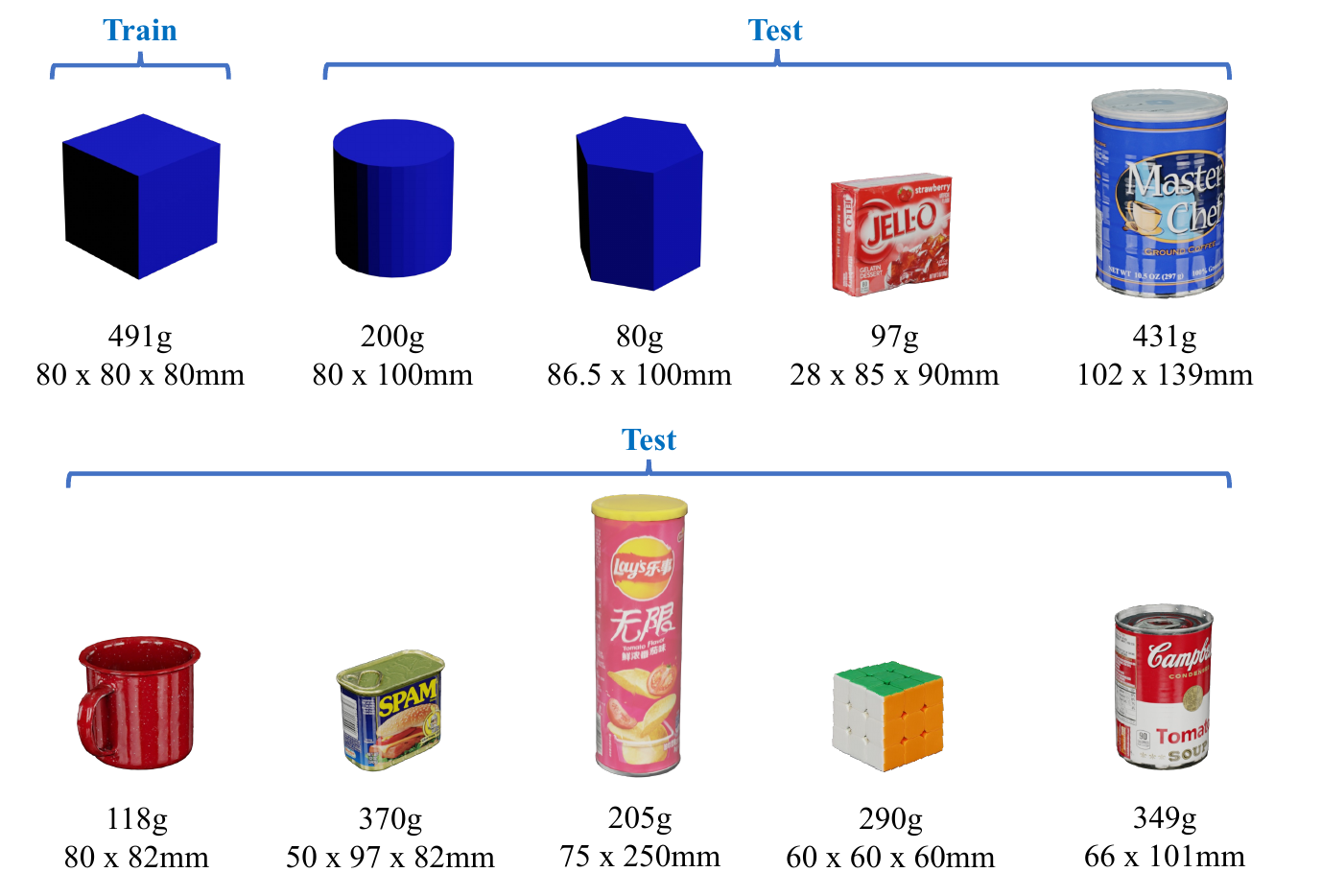}
     % \vspace{-1em}
    \caption{Objects used for training and testing in simulation. (Everyday objects are from the YCB Object Set \cite{calli2015benchmarking}.)}
  \label{fig:object_sets}
\end{figure}

To evaluate the generalizability of each RL agent, we examined the goal-reaching success rate in simulation for 10 objects (9 of which are unseen objects shown in Fig. \ref{fig:object_sets}), over 54 evenly distributed goals in a testing goal space shown in Fig. \ref{fig:training_workspace}a. We only considered goals that are at least 0.1\,m away from the starting location. The results are shown in Table \ref{table:objects_results}. All agents were able to generalize to the tested objects and goals. We found the most difficult objects to push were the gelatin box, which rotated much more easily than the cube due to its thin and long body, and the master chef can, which was large and therefore less easy to manipulate. Comparing each RL agent, the model-based agent performed the best for 9 out of 10 objects, demonstrating the strength of online planning when it comes to novel objects.

\begin{table}[h]
% \vspace{-1em}
\addtolength{\tabcolsep}{-1pt}
\caption{The success rate for each RL agent pushing 10 test objects over 54 evenly distributed goal locations across the workspace, averaged over 5 trials.}
\centering
\begin{tabular}{l|ccc} 
\hline
\multirow{2}{*}{\textbf{Objects\textbackslash{RL Agent}}}&  \multirow{2}{*}{\begin{tabular}[c]{@{}c@{}} \textbf{Model-Free} \\ \textbf{Tactile Image}\end{tabular}} & \multirow{2}{*}{\begin{tabular}[c]{@{}c@{}} \textbf{Model-Free} \\ \textbf{Tactile Pose}\end{tabular}}  &\multirow{2}{*}{\begin{tabular}[c]{@{}c@{}} \textbf{Model-Based} \\ \textbf{Tactile Pose}\end{tabular}}\\ 
&&&\\
\hline
\textbf{Cube}     & 0.91          & 0.89    & \textbf{0.96}\\ 
\textbf{Cylinder}         & 0.85         & \textbf{0.91}    & \textbf{0.91}\\
\textbf{Hexagonal Prism}  & 0.93 & 0.89  &  \textbf{0.94}\\ 
\textbf{Gelatin Box} & 0.71 & 0.74 & \textbf{0.75}\\
\textbf{Master Chef Can} & 0.57 & \textbf{0.76} & 0.66\\
\textbf{Mug} & 0.81 & 0.83 & \textbf{0.95} \\
\textbf{Potted Meat Can} & 0.80 & 0.89 &  \textbf{0.90}\\
\textbf{Chip Can} & 0.94 & 0.91 & \textbf{0.98}\\
\textbf{Rubiks Cube} & 0.96 & 0.94  & \textbf{0.99}\\
\textbf{Tomato Soup Can} & 0.89 & 0.89 & \textbf{0.94}\\
\hline
\end{tabular}
\label{table:objects_results}
\end{table}

% %--------------------------------------------------------    
\begin{figure*}[t]
  \centering
     \includegraphics[width=0.99\linewidth]{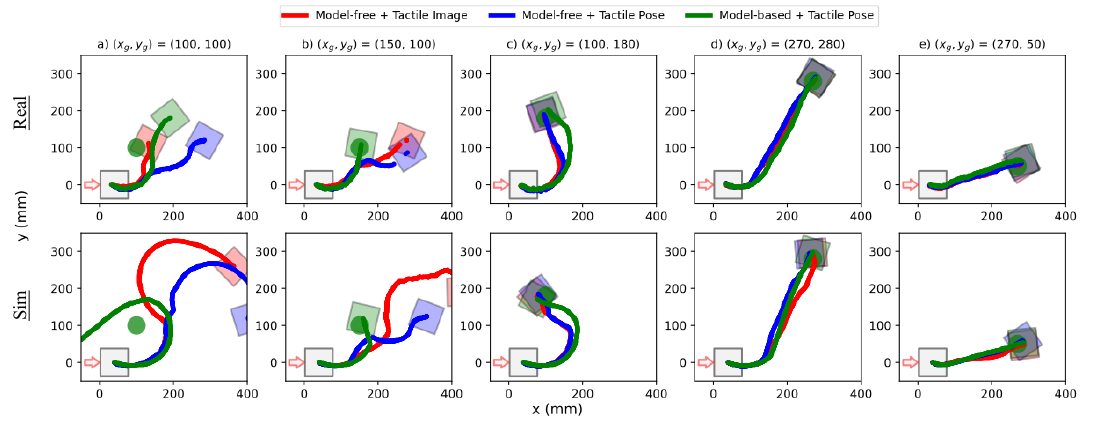}    \vspace{-1em}
    \caption{Pushing trajectories for 5 goal locations (denoted by green circular marker) in simulation and the real world. Each agent is intialized at the origin (red arrow) and terminates when the contact location is within 25\,mm of the goal.}
  \label{fig:objects_pushing_results}
  % \vspace{-0.5em}
\end{figure*}

% %--------------------------------------------------------    

  %---------------------------------------------------
\subsection{Real-world Performance}

In the physical experiments, we evaluate the RL agents against three performance criteria: 1) the ability to generalize to random goals within the workspace; 2) the ability to generalize to different objects with varying physical properties; and 3) the ability to be robust against disturbances.

\subsubsection{\textbf{Random Goals}}
We used the cube environment (see Fig.~\ref{fig:training_workspace}) to test the RL agents on 40 goals scattered evenly across the workspace. Pushing trajectories of the simulation and the real world are shown in Fig.~\ref{fig:objects_pushing_results}. 

For all considered RL agents, we achieved 100\% success rate for goals further than 110\,mm, demonstrating the reliability of the learned policies and validating our goal-conditioned tactile pushing formulation. There was a reduction in reliability for goals that were closer to the agent and positioned at large goal-bearing angles (Figs~\ref{fig:objects_pushing_results}a,b) because the agent was required to orientate the object sharply towards the goal, which can be challenging or even impossible given the action constraints described Sec \ref{subsec:rl_formulation}. %This is a limitation of the formulation. 

For each RL agent with goals placed at high bearing angles, the model-free agents outperformed the model-based RL agent (Figs \ref{fig:objects_pushing_results}c), taking a shorter path to the goal that corresponds to the higher rewards seen in Table~\ref{table:learning_results}. However, for more difficult goals closer to the agent, model-based RL produced more stable and interpretable pushing trajectories, whereas the model-free agents moved the object into uncontrollable regions. When comparing the results of the RL agents for different tactile observations, the similar final performance observed in both simulation and reality suggests that PoseNet also offers an accurate tactile sim-to-real observation model for contact surface pose. The minor discrepancy between the simulation and reality seen across all agents could originate from a mismatch of physical parameters such as friction. 

%-------------------------------------------------------- 
\begin{figure*}[t!]
  \centering
     \includegraphics[width=1.00\linewidth]{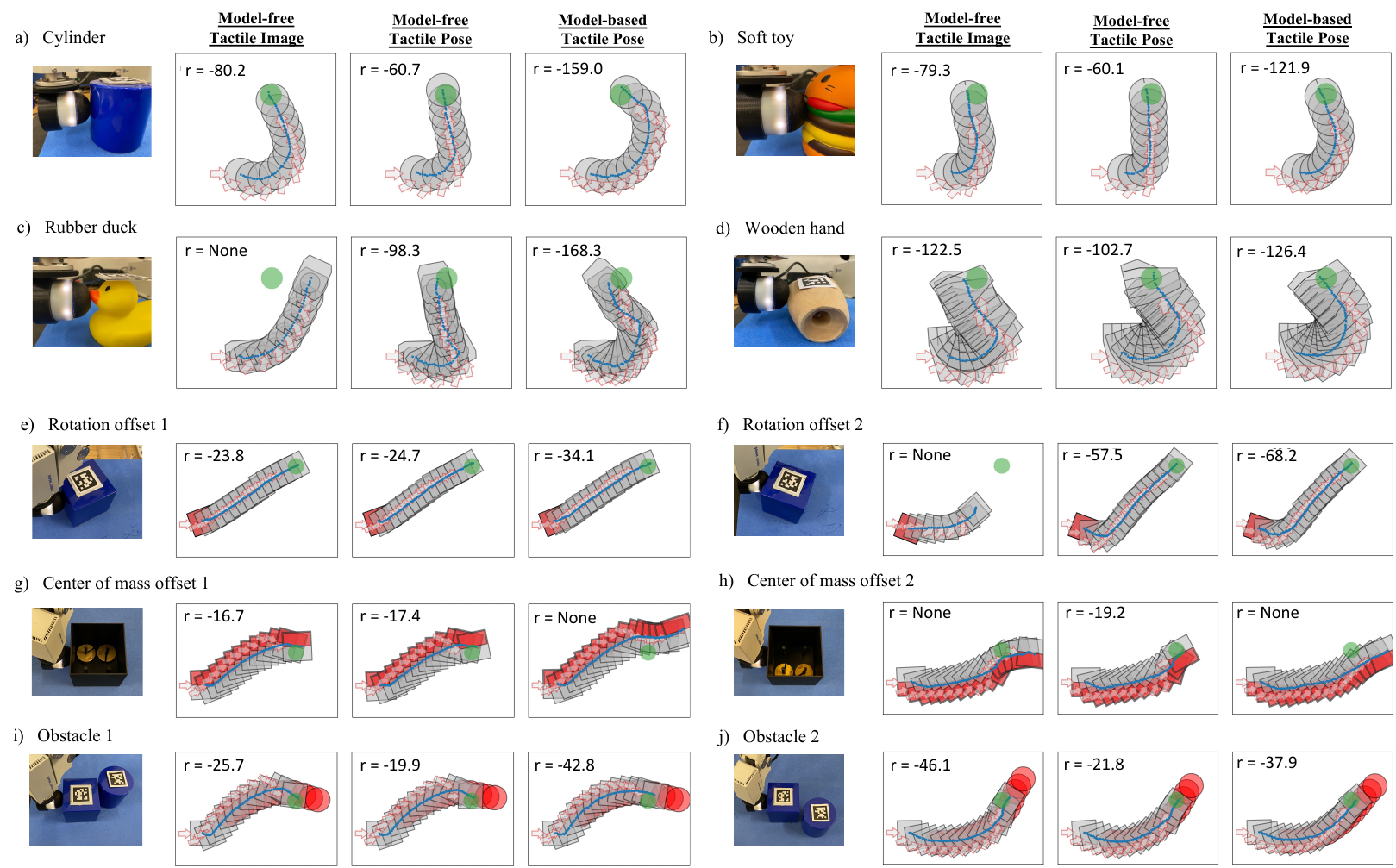}
    \vspace{0em}
    \caption{Sensor and object trajectories for various pushing scenarios. Red arrows represent the tactile sensor; blue dots represent the tracked object centers using ArUco markers; and green circles represent the goal regions. The episode reward (Eq. \ref{eq:push_reward}) is shown in each plot.}
  \label{fig:experiment_push_objects}
\end{figure*}
%-------------------------------------------------------- 

\subsubsection{\textbf{Irregular Objects}}
The ability to generalize to unseen objects is considered by pushing irregular objects with varying contact surfaces and physical properties. All objects are pushed from a stable configuration at the origin to a goal at $x=100$\,mm, $y=180$\,mm. Four objects were chosen such that: 1) a cylinder challenged the agent to maintain stable contact during large changes in pushing direction; 2) a soft deformable toy increased the uncertainties of the tactile signals, due to the compliant contact surface and the interaction can vary depending on the friction build up between the object and the pusher; 3) pushing a rubber duck on the beak produced an unstable pushing point and a contact deformation unseen during the sim-to-real training; and 4) finally a wooden hand had non-uniform contacts with the supporting surface which can make it quite sensitive to control (results in Fig.~\ref{fig:experiment_push_objects}a-d). 

When using tactile-pose-based observations, both model-free and model-based methods could generalize to all of the tested objects and successfully complete the task. The model-free agent took shorter paths than model-based RL, which is consistent with the results in the previous sections. The tactile-image-based model-free agent generally performed worse than the tactile-pose-based alternative: here it failed to push the duck to the goal, losing contact. Evidently, the RL agent was then unable to generalize to the unseen contact deformation. 

\subsubsection{\textbf{Disturbances}}
The robustness of the final policies is considered by applying various disturbances during pushing. First, we applied an initial contact angular offset of $\pm20^\circ$ (see Fig.~\ref{fig:experiment_push_objects}e,f). Next, a hollow box was pushed with 2$\times$100g weights on one side, corresponding to a center of mass offset (see Fig. \ref{fig:experiment_push_objects}g,h). Lastly, we placed an obstacle (cylinder weighing the same as the pushed object) in the path of the pushed object (see Fig.~\ref{fig:experiment_push_objects}\mbox{i,j}). 

% We placed the obstacle near the starting position where it would remain on the pushed object to become a coupled system. 

Once again, the tactile-pose-based RL agents were more reliable in all pushing scenarios, while the tactile-image-based agent failed with an initial contact angle offset of $-20^\circ$. This demonstrates limited generalizability when learning from tactile image observations. In the cases shown in Fig.~\ref{fig:experiment_push_objects}\mbox{g-j}, the presence of persistent disturbances made reaching the goal difficult for all agents, where a sticking contact restricted the movement of the pushing location along the contact surface to counteract a COF offset. However, the RL agents were still able to overcome these scenarios. Despite testing against large disturbances never seen during training, a policy learned using model-free RL still outperformed the online planner of model-based RL, consistently achieving shorter pushing paths. 

 %--------------------------------------------------------
%----------------------------------------------------------------------------------
\section{DISCUSSION AND FUTURE WORK} \label{sec:discussion}
% %----------------------------------------------------------------------------------
In this paper, we presented several successful deep RL approaches for solving the tactile pushing problem. These relied on a problem formulation that allowed us to obtain accurate and reliable policies for goal-conditioned pushing applicable to both model-free RL and model-based RL with online planning. These tactile-based agents were trained entirely in simulation and then transferred to reality using their respective observation models without additional training. 

Our experimental results showed that training with tactile-pose-based observations showed greater generalizability to unseen objects and increased robustness to disturbances. These results suggest that contact surface pose serves as a highly effective tactile feature for learning this task. In our comparison of different deep RL methods, we observed that model-based RL was more reliable for pushing a range of objects and goals, while requiring 100$\times$ fewer training data than other methods. However, with substantially more training data, model-free RL trained with tactile pose consistently obtained higher rewards across various pushing tests. This resulted in more stable and shorter pushing paths even in settings with large deviations from the training environment. In our view, this is due to our shaped reward, which rewards the agent for stable pushing, and a better final policy also resulted in greater robustness. Thus we expect that keeping the pusher normal to the contact surface is a strong stability criterion for object pushing, which then enables the generalization to a large variety of objects. 

Compared to the trajectory following problem explored in \cite{church_tactile_2021, lin2022tactile}, our policies can offer more general and versatile tactile pushing solutions for reaching an arbitrarily placed goal. Our problem formulation allows easy extension of the policies to reach a series of goals, which includes the trajectory following problem as a specific case. One limitation of our formulation is the action constraint, which restricts the reachability of goals close to the initial pushing location and limits exploration around the contact surface.  However, our results show that by sacrificing a small amount of control flexibility, there are significant benefits in the reliability of the learned policy. In the future, it would be interesting to remove the action constraint and further explore the trade-off between reliability and performance. Although here we leverage a sim-to-real RL approach, learning in the real world is feasible given the sample efficiency of model-based RL, which could offer additional performance gains. However, resetting the environment without human intervention and without visual information are practical challenges yet to be addressed. 
% We also demonstrated the advantages of a deep RL formulation compared with a state-feedback approach presented in \cite{lloyd2021goal} by overcoming known issues relating to the complex dynamics when approaching the goal. 
% We do believe online learning for tactile manipulation is an important problem for the future and encourage other researchers to take on this challenge. 
% Consequently, we believe that our approach is particularly valuable for object-pushing problems where the goals are broadly defined and the objective is a single abstract target. 

A key finding from this study was that RL policies trained without domain randomization or a diverse training curriculum that involved multiple objects were able to perform a wide range of pushing tasks, displaying strong generalization skills. This suggests that tactile information when used appropriately, can facilitate the efficient learning of general manipulation skills, and so make previously intractable robot manipulation problems tractable. Despite its widely-acknowledged value, tactile sensing remains an underutilized tool for robot learning. We hope that our work will inspire further research on using RL with tactile information and encourage its application to more challenging dexterous manipulation tasks.

{\em Acknowledgements:} We thank Haoran Li for his contribution to making CAD models. We thank Andrew Stinchcombe and Ugnius Bajarunas for helping with the 3D-printing of the stimuli and TacTips. We thank Georgia Chalvatzaki, Niklas Funk and Boris Belousov for the useful discussions. 

\bibliographystyle{unsrt}
\bibliography{manual}
% %------------------------------------------------------------------------------------------------------------------

\end{document}